\documentclass[lettersize,journal]{IEEEtran}
\usepackage{amsmath,amsfonts}
\usepackage{algorithmic}
\usepackage{array}
\usepackage[caption=false,font=normalsize,labelfont=sf,textfont=sf]{subfig}
\usepackage{textcomp}
\usepackage{stfloats}
\usepackage{url}
\usepackage{verbatim}
\usepackage{graphicx}
\def\BibTeX{{\rm B\kern-.05em{\sc i\kern-.025em b}\kern-.08em
    T\kern-.1667em\lower.7ex\hbox{E}\kern-.125emX}}
\usepackage{balance}

\usepackage{booktabs}
\usepackage{eccvabbrv}    %
\usepackage[accsupp]{axessibility}  %
\usepackage{array}
\usepackage{tabularx}
\newcolumntype{H}{>{\setbox0=\hbox\bgroup}c<{\egroup}@{}}
\usepackage{amsmath}
\usepackage{etoolbox}

\usepackage[dvipsnames]{xcolor}

\providetoggle{showcomments}
\settoggle{showcomments}{true} %

\iftoggle{showcomments}{%
    \newcommand{\resolved}[3][]{\ifstrequal{#1}{resolved}{\textcolor{blue}{RESOLVED:}~\textbf{{\MakeUppercase #2:}}~{#3}}{\textbf{\MakeUppercase #2:}~#3}}
    \newcommand{\jasper}[2][]{\textcolor{violet}{\resolved[#1]{jasper}{#2}}}
    \newcommand{\tm}[2][]{\textcolor{magenta}{\resolved[#1]{TM}{#2}}}
    \newcommand{\lluis}[2][]{\textcolor{RedOrange}{\resolved[#1]{lluis}{#2}}}
    \newcommand{\andre}[2][]{\textcolor{blue}{\resolved[#1]{andre}{#2}}}    
    \newcommand{\vitto}[2][]{\textcolor{OliveGreen}{\resolved[#1]{vitto}{#2}}}    
    \newcommand{\howard}[2][]{\textcolor{red}{\resolved[#1]{howard}{#2}}}    
    
}{%
    
    \newcommand{\jasper}[2][]{}
    \newcommand{\tm}[2][]{}
    \newcommand{\lluis}[2][]{}
    \newcommand{\andre}[2][]{}    
    \newcommand{\howard}[2][]{}    
    \newcommand{\vitto}[2][]{}    
}

\newcommand{\para}[1]{\vspace{2.5pt}\par\noindent\textbf{#1}}
\newcommand{\hammr}{HAMMR\xspace}

\usepackage{multirow}

\usepackage{alltt}
\usepackage{listings}
\usepackage[dvipsnames]{xcolor}
\usepackage[most]{tcolorbox}
\usepackage{courier}
\usepackage{chngcntr}
\usepackage{multicol}
\usepackage{capt-of}

\NewDocumentCommand\emojihammer{}{
    \includegraphics[scale=1.8]{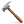}
}

\usepackage{hyperref}

\usepackage{orcidlink}

\begin{document}

\bstctlcite{IEEEexample:BSTcontrol}

\title{\hammr\emojihammer: HierArchical MultiModal React agents for generic VQA}
\author{Lluis Castrejon$^\star$,
Thomas Mensink$^\star$,
Howard Zhou$^\star$,
Vittorio Ferrari$^\dag$,
Andre Araujo$^\star$,
Jasper Uijlings$^\star$\\
\thanks{$^\star$Google DeepMind, $^\dag$Synthesia.io}
\thanks{Corresponding authors: \{lluisc, jrru\}@google.com.}
}

\maketitle

\begin{abstract}

Combining Large Language Models (LLMs) with external specialized tools (LLMs+tools) is a recent paradigm to solve multimodal tasks such as Visual Question Answering (VQA). While this approach was demonstrated to work well when optimized and evaluated for each individual benchmark, in practice it is crucial for the next generation of real-world AI systems to handle a broad range of multimodal problems.
Therefore we pose the VQA problem from a unified perspective and evaluate a single system on a varied suite of VQA tasks including counting, spatial reasoning, OCR-based reasoning, visual pointing, external knowledge, and more.
In this setting, we demonstrate that naively applying the LLM+tools approach using the combined set of all tools leads to poor results.
This motivates us to introduce \hammr: HierArchical MultiModal React. We start from a multimodal ReAct-based~\cite{yao23iclr_react} system and make it hierarchical by enabling our \hammr agents to call upon other specialized agents. This enhances the compositionality of the LLM+tools approach, which we show to be critical for obtaining high accuracy on generic VQA.
Concretely, on our generic VQA suite, \hammr outperforms a naive LLM+tools approach by 16.3\%.
Additionally, \hammr achieves state-of-the-art results on this task, outperforming
the generic standalone PaLI-X VQA model~\cite{chen23arxiv_palix} by 5.0\%.

\end{abstract}


\begin{figure*}[!t]
    \centering
    \includegraphics[width=\textwidth]{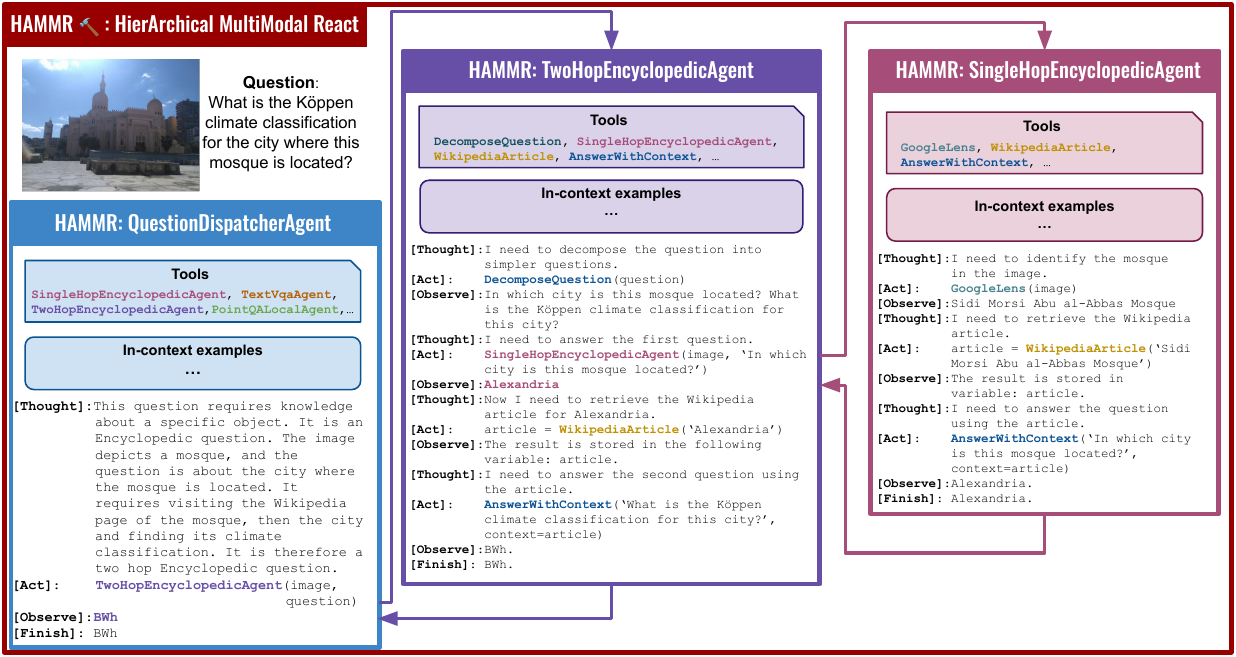}
    \caption{%
    {\bf Example of \hammr.} %
    \small We introduce HAMMR: HierArchical MultiModal React for generic visual question answering (VQA).
    Our work is an evolution of the LLM+tools approach~\cite{gupta23cvpr_visprog,hu23neurips_avis,suris23iccv_vipergpt}, where  we design a single system which can handle a large variety of VQA tasks.
    \hammr is a multimodal version of ReAct~\cite{yao23iclr_react} where agents themselves can act as tools. This results in a hierarchical and highly compositional approach where high-level \hammr agents can call lower-level agents dedicated to more specific tasks. This figure shows how HAMMR solves a two hop Encyclopedic-VQA~\cite{mensink23iccv_encvqa} question. Our high-level agent determines the question type and calls the corresponding encyclopedic two hop agent, which calls the single hop encyclopedic agent to solve the first part of the composite question. 
   }
    \label{fig:key_fig}
\end{figure*}

\section{Introduction}

Multimodal understanding and reasoning are key research problems in artificial intelligence (AI), which have witnessed substantial progress with the latest advancements in visual question answering (VQA) (\eg \cite{chen23arxiv_palix,gui22naacl,hu23cvpr,suris23iccv_vipergpt,gupta23cvpr_visprog,wang22tmlr_git}).
Today, the standard approach for VQA problems is to leverage Vision+Language Models (VLMs) \cite{openai2023gpt4,chen23arxiv_palix,chen23iclr_pali,alayrac22neurips}, which take an image with an associated textual question and are trained to produce a textual answer.
However, in the past year an alternative paradigm has been proposed: combine Large Language Models (LLMs) and computer vision tools to create flexible programs tailored to a given question \cite{suris23iccv_vipergpt,gupta23cvpr_visprog,hu23neurips_avis}.
This LLM+tools approach enables tackling new problems through in-context instructions instead of laborious and computationally expensive model finetuning, and facilitates incorporating retrieval mechanisms for better prediction of detailed knowledge where VLMs typically falter~\cite{chen23arxiv,mensink23iccv_encvqa}.

VQA problems have mostly been evaluated on individual benchmarks, each with specialized methods for specific types of questions (\eg~\cite{chen23iclr_pali,gui22naacl,gupta23cvpr_visprog,hu23neurips_avis,li20eccv_oscar,lu19neurips,suris23iccv_vipergpt,tan19emnlp_lxmert}). 
However, in practice it is crucial for the next generation of real-world AI systems to handle a broad range of multimodal questions without the user specifying what type of question they ask.
Therefore we pose the VQA problem from a unified perspective, aiming to build systems that can handle a wide variety of question types.
Our benchmarking covers many different VQA tasks, such as counting~\cite{acharya19aaai_tallyqa}, spatial reasoning~\cite{hudson19cvpr_gqa}, OCR-based reasoning~\cite{singh19cvpr_textvqa}, external knowledge~\cite{mensink23iccv_encvqa}, visual pointing~\cite{mani20arxiv_pointqa} and questions concerning two images~\cite{suhr19acl_nlvr2}. 
We explore the LLM+tools approach for this generic VQA setting for the first time -- in contrast to~\cite{suris23iccv_vipergpt,gupta23cvpr_visprog,hu23neurips_avis}, we want a single method which can solve any VQA question.
Our findings show that naively applying the LLM+tools approach in this generic setting by 
using the combined set of all tools leads to poor results.
This suggests that the large number of question patterns and different tools makes it difficult for the LLM to reason about which tools to combine for the specific question at hand.

In this work, we take a step forward in multimodal capabilities by proposing \textbf{\hammr} (HierArchical MultiModal React), a re-designed LLM+tools approach for generic VQA.
\hammr leverages a multimodal ReAct-based \cite{yao23iclr_react} system, where LLM agents can be prompted to select the most suitable tools to answer a given question.
Departing from previous work~\cite{gupta23cvpr_visprog,hu23neurips_avis,suris23iccv_vipergpt}, our agents may call upon other specialized agents which we construct for specific question types, effectively reusing specialized agents as tools.
This enhances the compositionality of the LLM+tools approach by enabling reuse of tool combinations that are effective for complex tasks (Fig.~\ref{fig:key_fig}).
Careful experimentation on our generic VQA suite shows that this hierarchical setup is critical for achieving high performance in this generic setting, outperforming naive extensions of existing LLM+tools methods by $16.3\%$.
Additionally, \hammr achieves state-of-the-art results, outperforming the recent PaLI-X VLM \cite{chen23arxiv_palix} by $5.0\%$.

In summary, 
we address a generic VQA problem, which is more applicable and scalable for real-world applications -- contrasting to a conventional specialized VQA setup, where existing datasets typically focus only on a specific question type~\cite{acharya19aaai_tallyqa,mani20arxiv_pointqa,mensink23iccv_encvqa}.
To that end we combine $6$ datasets spanning $8$ diverse question types to build a comprehensive evaluation suite.
We propose \hammr, a hierarchical LLM+tools approach to address generic VQA.
\hammr is based on multimodal ReAct which allows specialized agents to be reused as tools, enabling a compositional design which facilitates agent development.
Finally, we perform comprehensive experiments on our generic VQA evaluation suite, demonstrating state-of-the-art performance, with a $16.3\%$ gain over naively extending LLM+tools approaches and a $5.0\%$ improvement over PaLI-X. 
    
\section{Related Work}

\para{VQA models with tools.} Various recent works in VQA propose retrieval-augmented models ~\cite{gui22naacl,hu23cvpr,lin22neurips,marino21cvpr,mensink23iccv_encvqa,vickers21acl}. These models leverage existing knowledge bases of commonsense knowledge~\cite{marino21cvpr}, external 
knowledge~\cite{gui22naacl,hu23cvpr,lin22neurips,marino21cvpr,mensink23iccv_encvqa,vickers21acl}, or knowledge of object relations~\cite{marino21cvpr,hu23cvpr}. Knowledge can be represented as triplets~\cite{gui22naacl,hu23cvpr,lin22neurips,marino21cvpr,vickers21acl}, image-text pairs~\cite{hu23cvpr}, or free-form text~\cite{mensink23iccv_encvqa}.
In KRISP~\cite{marino21cvpr} and REVEAL~\cite{hu23cvpr} they train the retrieval component. In KAT~\cite{gui22naacl}, InFactuality~\cite{vickers21acl}, and REVIVE~\cite{lin22neurips} they keep this component fixed. 
In Encyclo\-pedic-VQA~\cite{mensink23iccv_encvqa} they use Google Lens as an external retrieval tool.

Complementary, several works show that LLMs can solve VQA tasks without pre-training by translating the image into text which is given as context to the question asked. In particular, PICa~\cite{yang22aaai} converts images to captions (using VinVL~\cite{zhang21cvpr}) and feeds this to GPT-3 with a few in-context examples.
PromptCap~\cite{hu23iccv} improves on this by training a question-dependent caption generator, encouraging captions to contain the required information. 
For Encyclopedic-VQA~\cite{mensink23iccv_encvqa} fine-grained knowledge is required. So their method is to identify the specific entity depicted, retrieve its Wikipedia page, and feed this as context along with the question to PaLM~\cite{chowdhery22arxiv} or GPT-3~\cite{brown20neurips}.

More recently, VisProg~\cite{gupta23cvpr_visprog} and ViperGPT~\cite{suris23iccv_vipergpt} leverage the in-context learning capabilities of LLMs in combination with tool-use to solve complex visual tasks. Thereby they bypass finetuning any model. To solve a visual task, the LLM receives the task instructions and generates a program which calls specific tools in a certain order. Some of these tools inspect the image (e.g. classification, detection, VQA).
Both~\cite{gupta23cvpr_visprog,suris23iccv_vipergpt} demonstrate they can solve a large variety of tasks. However, each task is addressed separately by creating task specific in-context examples while restricting the available tools. In this paper we explore how we can create a single system which solves a large variety of VQA tasks.

AVIS~\cite{hu23neurips_avis} also addresses VQA using LLMs with in-context learning and tool-use. While~\cite{gupta23cvpr_visprog,suris23iccv_vipergpt} generate a complete program which is then executed, AVIS calls tools in an iterative fashion. As such, AVIS can backtrack from unpromising steps and recover from errors. Similar to~\cite{hu23neurips_avis}, our system
uses iterative decision making but we use the ReACT paradigm~\cite{yao23iclr_react}. In contrast to~\cite{hu23neurips_avis}, we explore how to create a single agent which can solve a large variety of VQA questions.

Concurrent to our work,~\cite{yang23arxiv_mmreact} is a visual demo of a multimodal ReAct implementation lacking any quantitative evaluation. 
In our paper we introduce a hierarchical version of multimodal ReAct. In addition, we also provide a proper quantitative evaluation on a suite of varied VQA datasets.

\para{Planning and tool-use in LLMs.}
With the advent of large language models~\cite{brown20neurips,chowdhery22arxiv} it has become possible to forego finetuning and solve a new task by in-context learing~\cite{brown20neurips}.
This works particularly well by inviting the model to reason, leading to increasingly complex planning strategies: Ranging from `think step-by-step'~\cite{kojima22neurips}, to Chain-of-Thought~\cite{wei22neurips,zhou23iclr}, to various tree- or graph-based planning algorithms~\cite{besta23arxiv_graphofthoughts,yao23treeofthoughts,hao23emnlp_rap,xie23neurips_beamsearchllm}.

Orthogonally, various works augment LLMs with external tools for improving aspects like factuality~\cite{borgeaud22icml,guu20icml,khattab23arxiv_dsp,schick23arxiv_toolformer}, math abilities~\cite{schick23arxiv_toolformer,parisi22arxiv_talm}, etc. Notably, ReAct~\cite{yao23iclr_react} and ART~\cite{paranjape23arxiv_art} combine iterative planning with tool use. In this paper we start from ReAct~\cite{yao23iclr_react} and make it both multimodal and hierarchical.

\para{Traditional VQA models.}
VQA models are typically neural networks which take as input an image and a question and are trained in a discriminative way to produce the correct answer~\cite{li20eccv_oscar,lu19neurips,tan19emnlp_lxmert}. More recently, a generation of generative multimodal models is trained on a mixture of different tasks, resulting in flexible models which can follow in-context instructions~\cite{alayrac22neurips,chen23iclr_pali,chen23arxiv_palix,hao22arxiv,li23iclm_blip2}. Since our aim is to have a single model solving a variety of VQA tasks, we perform a comprehensive comparison to PaLI-X~\cite{chen23arxiv_palix} for generic VQA.

\para{Generic VQA.} The popular VQA~\cite{antol15iccv} and VQAv2~\cite{goyal17cvpr} datasets were specifically designed to have open-ended questions which require a wide range of capabilities to solve. However, many types of questions were not covered, leading to the introduction a variety of specialized datasets~\cite{acharya19aaai_tallyqa,chen23arxiv,hudson19cvpr_gqa,mani20arxiv_pointqa,marino19cvpr_okva,mensink23iccv_encvqa,schwenk22eccv_aokvqa,shah19aaai,singh19cvpr_textvqa,suhr19acl_nlvr2}. In this paper, we evaluate a single system on a suite of modern specialized VQA datasets, going back to the original intent of~\cite{antol15iccv,goyal17cvpr}, but covering a more diverse collection of question types.
    
\section{Method}\label{sec:method}

\newcommand{\taoexample}[3]{%
\vspace{1mm}{}\noindent{\footnotesize
\hspace{1mm}
\begin{tabular}{@{}p{.20\columnwidth}@{}p{.75\columnwidth}@{}}
    \textbf{\texttt{[Thought]:}}&\texttt{#1}\\
    \textbf{\texttt{[Act]:}}&\texttt{#2}\\
    \textbf{\texttt{[Observe]:}}&\texttt{#3}
\end{tabular}
\vspace{1mm}
}}

\subsection{Preliminaries: ReAct}\label{sec:method_react}
ReAct~\cite{yao23iclr_react} is an agent which interleaves verbal reasoning with actions and observations to make interactive decisions. For its reasoning it uses a frozen, decoder-only LLM with chain-of-thought prompting~\cite{wei22neurips}.
We call this LLM which does the high-level reasoning the \emph{orchestrator LLM}.
Actions are initiated when the orchestrator LLM generates specific instructions and their results are fed back into the orchestrator as part of the prompt.
In this paper an action calls an external tool and receives its output.

ReAct agents perform three steps: a thought to plan what to do next, an act calling a tool, and observe the output returned by the tool. For example: 

\taoexample{I need to search for Alexandria and find information about \ldots}{Search(Alexandria)}{Alexandria is a Mediterranean port city in Egypt \ldots}

The \texttt{[Thought]} is generated by the orchestrator LLM and formulates what should be done next in natural language.
The \texttt{[Act]} specifies in a line of code the tool call with the exact input arguments. This is also generated by the orchestrator LLM and should logically follow from the previous \texttt{[Thought]}.

Next, the tool is invoked and its observation is fed back in plain text to the orchestrator after \texttt{[Observe]}.
After these three steps, the ReAct agent can decide to \texttt{[Finish]} and provide an answer, or to start another \texttt{[Thought]}-\texttt{[Act]}-\texttt{[Observe]} iteration.
As each iteration continuously grows the prompt, the ReAct agent has access to all previous reasoning steps, tool calls, and observations when planning the next step.

The original ReAct~\cite{yao23iclr_react} is purely text based: the arguments of all tool calls are text produced by the orchestrator LLM, and all tools produce observations consisting of plain text.
ReAct cannot handle other modalities such as images.

\subsection{Multimodal ReAct for generic VQA}\label{sec:method_generic_react}

\para{Multimodality through processing variables.}
To make ReAct multimodal, we give it access to variables, which could contain images, text, or other data types.
In the prompt we provide in-context examples of function calls that input and return variables in an \texttt{[Act]} step.
The returned variable is stored, and its name is mentioned in the \texttt{[Observe]} step, so that it is available for further reasoning and actions.
The input image of the VQA question is stored beforehand in a special `image' variable, so the agent can start the process.
For example:

\taoexample{I need to crop the top left corner of the image, to detect \ldots}{crop = CropImage(image, [0,0,50,50])}{Output of `CropImage` is stored in the variable: `crop`}

Hence, subsequent tool calls can access the cropped image data stored in the \texttt{crop} variable
(see also Fig.~\ref{fig:pointqa_specialists}).
Other tools could return a variety of data types, including bounding-boxes, IoU scores, or a whole Wikipedia page (Tab. \ref{tab:tools}). In all cases, the tool output is stored in a variable that the agent can reuse.

\begin{figure*}[th]
    \centering
    \includegraphics[width=\linewidth]{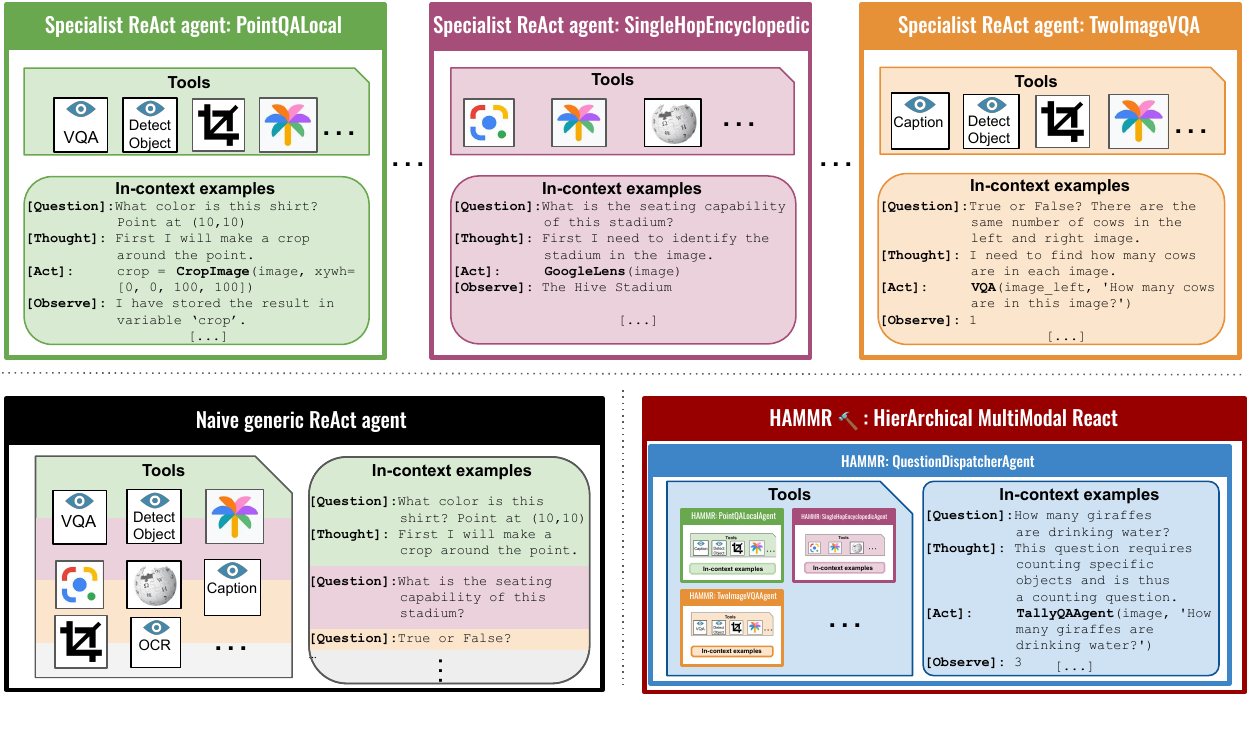}
    \vspace{-.4cm}
    \caption{{\bf Illustration of our approach}. 
    \emph{Top}: the common approach~\cite{hu23neurips_avis,gupta23cvpr_visprog,suris23iccv_vipergpt} is to create a specialist orchestrator agent for each individual task.  
    \emph{Bottom left}: To create a generalist orchestrator agent, the straightforward approach is to collect all tool descriptions and all in-context examples of each individual specialist.
    \emph{Bottom right}: We propose \hammr: we allow ReAct agents themselves to be called as tools. This leads to a hierarchical and modular approach where high-level ReAct agents can call agents which have a more specific task. For our generic VQA setting, our high-level agent determines the type of VQA question, after which it dispatches the question to the corresponding specialist HAMMR agent.
    }
    \label{fig:illustration_methods}
    \vspace{-.4cm}
\end{figure*}

\setlength{\tabcolsep}{6pt}
\begin{table}[t]
    \vspace{-.2cm}
    \caption{List of tools used by our agents.}
    \label{tab:tools}
    \centering
    \resizebox{\linewidth}{!}{
        \begin{tabular}{llll}
        \toprule
        {\bf Tool} & {\bf Input} & {\bf Output} & {\bf Backbone} \\
        \midrule
        GoogleLens           & image                  & entity            & Google Lens   \\
        DetectObject         & image + class name     & boxes             & OwlViTv2      \\
        ObjectInImage        & image + class name     & yes/no            & BLIP-2/PaLI-X \\
        OCR                  & image                  & recognized text   & BLIP-2/PaLI-X \\
        Caption              & image                  & caption           & BLIP-2/PaLI-X \\
        VQA                  & image + question       & answer            & BLIP-2/PaLI-X \\
        CropImage            & image + box            & image             & PIL           \\        
        SpatialSelection     & boxes + relation       & box               & NumPy         \\
        BoundingBoxOverlap   & anchor box + $n$ boxes & $n$ IoU scores    & NumPy         \\ 
        WikipediaArticle     & entity                 & Wikipedia page    & Wikipedia     \\
        AnswerWithContext    & question + context     & answer            & PaLM 2        \\
        DecomposeQuestion    & question               & two questions     & PaLM 2        \\
        \bottomrule
        \end{tabular}
        }  %
        \vspace{-.3cm}
\end{table}

\para{Tools for VQA.}
For solving VQA tasks we need tools which operate on images (Tab.~\ref{tab:tools}).
We provide fine-grained entity recognition with Google Lens~\cite{googlelens},
Object Detection using OwlViTv2~\cite{minderer22eccv_owlvit, owlvit2},
and OCR, Captioning, and ObjectInImage using either BLIP-2~\cite{li23iclm_blip2} or PaLI-X~\cite{chen23arxiv_palix}.
Additionally, analog to~\cite{gupta23cvpr_visprog} we also have VQA itself as a tool through these VLMs, which is used to ask specific questions about the image or an image region cropped by the orchestrator during its reasoning.

Next to the image analysis tools, we have
a simple crop image tool,
several tools dealing with bounding boxes,
a tool for retrieving a Wikipedia page,
and two tools which use an LLM to perform specialized functions on text (answer a question given some context, and decomposing a question into two simpler ones; both used in Fig.~\ref{fig:key_fig}).

\para{Naive generic ReAct agent.}
A ReAct agent for a benchmark with a certain question type is constructed by specifying a prompt with a list of tool descriptions and in-context examples showing how to solve a few questions by using those tools.
Hence the straightforward way to make a generic ReAct agent working on a broad set of question types is to create a long prompt listing all tools and all in-context examples across all types.
Such generic ReAct agent is visualized in Fig.~\ref{fig:illustration_methods} and acts as our main baseline, representing the most direct extension of current LLM + Tools approaches to generic VQA.

\subsection{\hammr: HierArchical MultiModal React}\label{sec:method_hammr}

The naive generic ReAct agent has all the necessary information to solve the generic VQA task.
However, as the number of question types increases, the number of tools and in-context examples required leads to a very long LLM prompt with a large number of reasoning patterns.
It is unclear if the orchestrator LLM can correctly attend to the relevant parts of such a long prompt, reusing the right (portion of an) example suitable for a new input question.
Furthermore, if the orchestrator starts making planning mistakes, their cause is hard to identify, making the system difficult to debug.

Therefore, we introduce \hammr: HierArchical MultiModal React. 
To answer a broad range of visual question types, we enable agents to call upon other specialized agents which are focused on a specific question type -- each specialized agent can be reused as a tool.
This leads to a compositional approach which enables solving increasingly complex tasks, while limiting the complexity that each individual agent needs to handle.
While the generic ReAct agent requires a long prompt specifying many tools and examples all at the same time,
\hammr modularizes generic reasoning by relying on specialized agents, each requiring a much smaller prompt involving a single reasoning pattern and a small number of tools.
This makes the task of each agent in \hammr simpler, since the solution to the problem is distributed across agents at different reasoning levels.

We start by creating a specialized agent for each VQA question type.
For example, questions requiring commonsense reasoning may leverage VLMs by using a tool such as `Caption', while questions requiring external knowledge may leverage Google Lens and Wikipedia.
Each specialized agent is created via prompts which are refined via iterations on a validation set, without any training of the orchestrator LLM parameters.
For some question types, there may exist multiple valid reasoning patterns involving different tools.
In this case, we experiment with each variant on a small validation set and pick the one with highest accuracy.
Specialized agents may also reuse other specialized agents as part of their reasoning chain.
For example, two-hop questions may require single-hop reasoning, where a single-hop agent could be reused (Fig.~\ref{fig:key_fig}).

To tackle generic VQA, we then create a high-level dispatcher agent which first determines the type of question and then calls the appropriate specialist HAMMR agent.
Our question dispatcher agent is also created via prompts, which we refine on a small validation set consisting of a few dozen examples for each question type (see Tab.~\ref{tab:dataset_overview}). To quickly iterate we only verify whether the correct specialist agent is identified without actually invoking it as a tool, highlighting how the compositionality of HAMMR speeds up development.

We visualize our approach in Fig~\ref{fig:illustration_methods} (bottom-right) and show an example of how it solves a VQA question hierarchically in Fig.~\ref{fig:key_fig}, where 3 reasoning levels are used.

Our design with a high-level dispatcher and specialized agents has multiple benefits over the naive generic agent:
(1) Agents are task-specific, which enables researchers to focus on a single problem at a time;
(2) Improving a task-specific agent will lead to improvement of the overall system, because it will more often solve that task successfully when invoked. In contrast, for a generic ReAct agent, progress on one task may mean a regression on another, since the reasoning process for all tasks is entangled.
(3) The compositionality of our approach means it is easier to debug since failure modes can be attributed to specific agents.
    
\section{Results}\label{sec:results}

\subsection{Experimental setup}

\begin{figure*}[t]
    \centering
    \includegraphics[width=\linewidth]{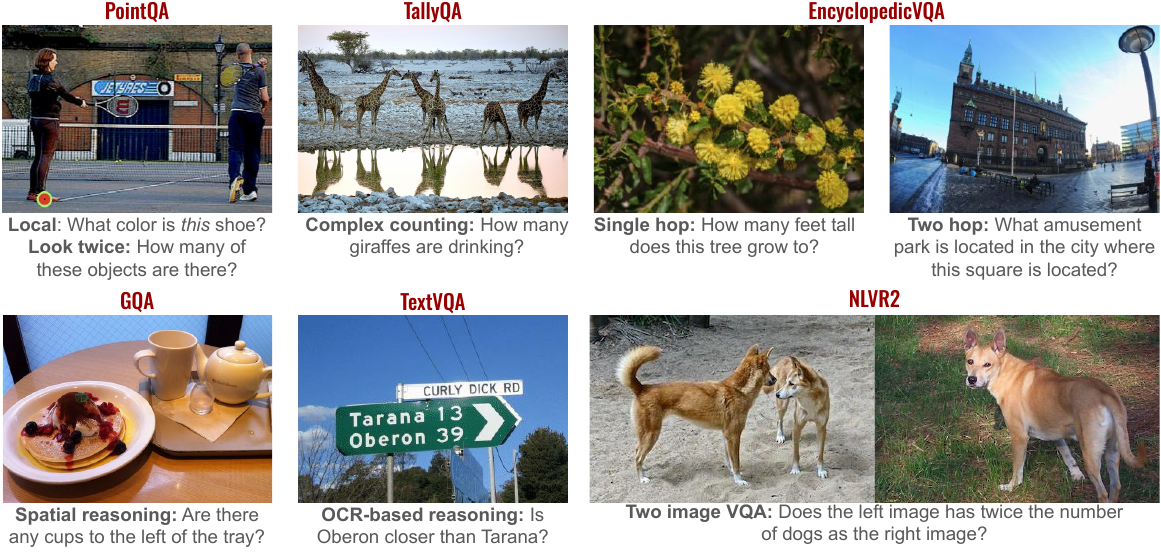}
    \caption{
    {\bf Question type examples.} 
    Example images and questions from the six datasets spanning 8 question types used to evaluate \hammr.
    }
    \label{fig:dataset_examples}
\end{figure*}

\setlength{\tabcolsep}{3pt}
\begin{table*}[t]
    \caption{
    \textbf{Generic VQA setup.}
    Overview of the datasets, question types, evaluation metrics and available tools.
    }
    \label{tab:dataset_overview}
    \centering    
    \resizebox{0.65\linewidth}{!}{
    \begin{tabular}{lllc} 
        \toprule
        \textbf{Dataset} & \textbf{Question Type} & \textbf{Available Tools} & \textbf{Metric} \\
        \toprule
        \multirow{2}{*}{PointQA~\cite{mani20arxiv_pointqa}}
            & Local pointing  & CropImage, VQA, DetectObjects & \multirow{2}{*}{Exact Match} \\
            & Look twice      & CropImage, VQA, Caption & \\
            \cmidrule{1-4}
        \multirow{3}{*}{EncVQA~\cite{mensink23iccv_encvqa}} 
            & Single hop & GoogleLens, WikipediaArticle, AnswerWithContext & \multirow{3}{*}{BERT Matching} \\
            & \multirow{2}{*}{Two hop} & GoogleLens, WikipediaArticle, AnswerWithContext,  & \\
            & & DecomposeQuestion, SingleHopEncyclopedicAgent \\ 
            \cmidrule{1-4}
        NLVR2~\cite{suhr19acl_nlvr2}
            & Two image VQA & VQA & Exact Match \\
            \cmidrule{1-4}
        \multirow{2}{*}{GQA~\cite{hudson19cvpr_gqa}}
            & \multirow{2}{*}{Spatial reasoning} & DetectObject, CropImage, ObjectInImage, OCR, & \multirow{2}{*}{Exact Match} \\
            & & BoundingBoxOverlap, SpatialSelection & \\
            \cmidrule{1-4}
        TallyQA~\cite{acharya19aaai_tallyqa}
            & Complex counting & DetectObject, CropImage, ObjectInImage & Exact Match \\
            \cmidrule{1-4}
        TextVQA~\cite{singh19cvpr_textvqa}
            & OCR-based reasoning & DetectObject, CropImage, OCR & VQA Accuracy \\
        \bottomrule
    \end{tabular}
    }
\end{table*}

\para{Datasets.}
In order to evaluate the capabilities on generic VQA, we select multiple VQA datasets, each focusing on a specific question type (Tab.~\ref{tab:dataset_overview}, Fig.~\ref{fig:dataset_examples}):
\textbf{\textit{PointQA}}~\cite{mani20arxiv_pointqa} questions include a reference point on the image, simulating the pointing gesture. 
We use this dataset for two question types: {\em local pointing} which asks a question about the point, and {\em look twice} which requires understanding the local context around the point as well as the whole image.
\textbf{\textit{EncyclopedicVQA}}~\cite{mensink23iccv_encvqa} features visual questions about detailed properties of fine-grained categories and instances.
We use this dataset for {\em single hop} encyclopedic questions, where the answer can be found on the Wikipedia page of the corresponding class/instance. 
We also use this dataset for {\em two hop} questions, which requires accessing two Wikipedia pages to obtain the final answer.
In \textbf{\textit{NLVR2}}~\cite{suhr19acl_nlvr2} two images are shown along with a statement, and the question is whether the statement is true or false ({\em two image VQA}).
\textbf{\textit{TallyQA}}~\cite{acharya19aaai_tallyqa} offers {\em complex counting} questions, that involve an additional reasoning / identification step.
\textbf{\textit{TextVQA}}~\cite{singh19cvpr_textvqa} questions require reading and reasoning about text in the image ({\em OCR-based reasoning}).
The final dataset we include is
\textbf{\textit{GQA}}~\cite{hudson19cvpr_gqa}, which mostly contains questions involving reasoning about spatial relations between objects ({\em spatial reasoning}). %

\para{Evaluation.}
To limit inference costs of the orchestration approach we follow~\cite{gupta23cvpr_visprog,suris23iccv_vipergpt} and evaluate on a subset of the data. From each dataset we select 1000 random samples from the test set (or validation if no test set is available).
We measure either exact match (EM) accuracy, VQA accuracy~\cite{antol15iccv}, or BERT Matching accuracy~\cite{bulian22tomayto, mensink23iccv_encvqa} depending on the metric proposed by the dataset authors (Tab. \ref{tab:dataset_overview}).
The final metric is the average of the per-dataset accuracies.
For models or agents using in-context examples, we select samples from the training sets of the datasets and manually annotate them ourselves.

\para{Implementation details.}
As orchestrator we use PaLM 2~\cite{chowdhery22arxiv} for text. It is publicly available via the PaLM API as \texttt{text-bison@001}~\cite{palm_api}. It was fine-tuned to follow natural language instructions on a variety of language tasks and further fine-tuned using RLHF~\cite{ouyang22neurips_rlhf}.
It supports a prompt of 8192 input tokens and can generate up to 1024 output tokens.
In most of our experiments we use the publicly available BLIP-2 T5-XXL model~\cite{li23iclm_blip2} for asking questions about the image. When comparing to state-of-the-art VLMs in Sec.~\ref{sec:results_palix} we replace this model with PaLI-X 55B multitask VQA finetuning~\cite{chen23arxiv_palix}. For object detection we use the publicly available OWL-ViTv2 CLIP L/14 ST+FT model~\cite{minderer2023neurips_owlvit2}.
    
\subsection{LLM+Tools for generic VQA}\label{sec:results_llm_tools}

\setlength{\tabcolsep}{6pt}
\begin{table}[t]
    \caption{
    \textbf{Results: Specialists vs Naive Generic ReAct vs HAMMR.} The specialist agents serve as an approximate upper bound.
    }
    
    \label{tab:results_llm_tools}
    \centering
    \resizebox{\linewidth}{!}{
    \begin{tabular}{l|c|cc}
        \toprule
        Dataset             & Specialist ReAct Agents & Naive Generic React & HAMMR \\
        \midrule
        PointQA local       & 48.1 & 8.7   & \bf{47.8} \\
        PointQA look twice  & 55.0 & 46.1  & \bf{55.0} \\
        EncVQA single hop   & 51.8 & 9.8   & \bf{45.0} \\
        EncVQA two hop      & 25.9 & 13.9  & \bf{22.8} \\
        NLVR2               & 61.0 & 37.2  & \bf{55.4} \\
        GQA                 & 50.7 & 41.5  & \bf{50.5} \\ 
        TallyQA             & 28.6 & 25.4  & \bf{29.2} \\
        TextVQA             & 22.0 & 9.7   & \bf{16.3} \\
        \midrule
        Average performance & 42.9 & 24.0  & \bf{40.3} \\
        \bottomrule
    \end{tabular}
    }  %
\end{table}

\para{Specialist ReAct agents.}
First we develop and evaluate specialist ReAct agents for each question type and dataset separately, in line with previous LLM+Tools works~\cite{hu23neurips_avis,gupta23cvpr_visprog,suris23iccv_vipergpt}.
Later we will reuse the prompts of the specialist ReAct agents or the agents themselves for the generic VQA setting.
These specialist agents serve as an approximate upper-bound for the generic VQA problem, since we develop and evaluate each specialist on its own specific type of VQA question.
This corresponds to knowing the question type at test time, giving a large information advantage over the generic VQA setting addressed later in this section.
Tab.~\ref{tab:dataset_overview} shows the available tools for each of our specialist ReAct agents.

Tab.~\ref{tab:results_llm_tools}, left column, shows the results of evaluating each specialist agent on its own dataset.
To put these numbers in context, on NLVR2 VisProg~\cite{gupta23cvpr_visprog} reports 62.4\%, vs 61.0\% for our specialist agents.
On GQA we report $50.7\%$ whereas VisProg~\cite{gupta23cvpr_visprog} reports 50.5\% and ViperGPT~\cite{suris23iccv_vipergpt} reports 48.1\%.
These numbers are not directly comparable since all works use slightly different components: VisProg uses GPT3~\cite{brown20neurips} as orchestrator, OWL-ViT~\cite{minderer22eccv_owlvit} as object detector, and ViLT~\cite{kim21icml_vilt} for VQA.
ViperGPT uses GPT3-Codex~\cite{chen21arxiv_gpt3codex} as orchestrator, GLIP~\cite{li22cvpr_glip} as object detector, and BLIP-2 T5-XXL~\cite{li23iclm_blip2} for VQA.
Nevertheless, we conclude that our specialist agents form a solid multimodal ReAct-style implementation of the LLM+Tools approach.
We will use them within our HAMMR approach below.

\para{Naive generic ReAct agent.}
We build a generic ReAct agent by collecting all tool descriptions and in-context examples from all specialists agents into a single long prompt (Sec.~\ref{sec:method_generic_react}, Fig.~\ref{fig:illustration_methods} bottom-left).
Results in Tab.~\ref{tab:results_llm_tools} show that over the whole benchmark the generic ReAct agent performs quite poorly (24.0\%). This is much lower than the approximate upper-bound of $42.9\%$ of applying the specialist agents on their own datasets (Tab.~\ref{tab:dataset_overview}).
This confirms our hypothesis that the naive generic agent starts confusing the in-context examples for different question types (Sec.~\ref{sec:method_hammr}).
We give a more detailed analysis in Sec.~\ref{sec:analysis}.

\para{\hammr.}
To address generic VQA using \hammr, we implement a high-level `question dispatcher' agent which determines the type of question and dispatches it to the appropriate specialist (Sec.~\ref{sec:method_hammr}, Fig~\ref{fig:illustration_methods} bottom-right).
Furthermore, \hammr enables specialist agents to call other specialist agents as in Fig.~\ref{fig:key_fig}.
Results in Tab.~\ref{tab:results_llm_tools} show that \hammr outperforms the naive generic agent by a margin of $16.3\%$ ($40.3\%$ vs $24.0\%$).
This demonstrates that our hierarchical and compositional approach is superior for generic VQA.
Furthermore, HAMMR performs quite close to the approximate upper-bound of the specialist agents (40.3\% vs 42.9\%).

\begin{figure}[t]
    \centering
    \resizebox{\linewidth}{!}{
    \includegraphics[height=55mm]{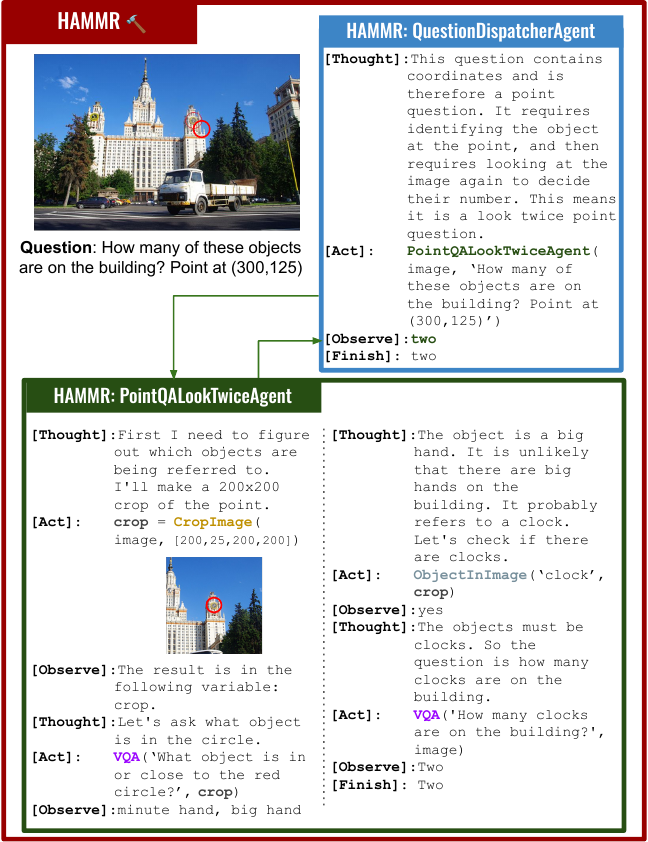}
    }
    \caption{{\bf Example of specialist HAMMR agents solving PointQA LookTwice} To solve this question, the agent first takes a crop of the image to focus on the object of interest. Interestingly, the LLM orchestrator correctly identifies that "minute hand, big hand" refers to a clock, recovering from an imperfect response of the VQA tool.
    }
    \label{fig:pointqa_specialists}
\end{figure}

Let's now consider how \hammr successfully solves a few questions. In Fig.~\ref{fig:key_fig} the question dispatcher agent calls the two hop encyclopedic agent, which in turn calls the one hop agent. This keeps the input prompt of the two hop agent focused and more easy to reason with.
Furthermore, while the retrieved Wikipedia article consists of text, we found that putting it in a variable and having a separate LLM call to answer the question avoids clogging the prompt of the agent.
Hence, adding variables to a ReAct agent not only enables it to be multi-modal, but can also help with text-only subproblems.
On PointQA look twice (Fig.~\ref{fig:pointqa_specialists}), the question dispatcher successfully identifies that it has to look at the image twice: first identify the object under the point input and then perform a counting task.
Thus the dispatcher calls the PointQALookTwice agent, which uses image crops to better identify the object of interest. Interestingly, the orchestrator LLM within the PointQALookTwice agent succesfully maps the identified object part (minute hand, big hand) to the object of interest (clock, see third Thought).
    
\subsection{Error Analysis}\label{sec:analysis}

\para{Setup.}
We analyze the errors made by the specialist agents, the naive generic agent, and \hammr.
We categorize three types of errors which roughly align with the key ReAct steps (Thought/Act/Observe, Sec.~\ref{sec:method_react}):

{\emph{Reasoning}}. The orchestrator LLM makes a mistake in its reasoning process. This manifests itself typically in the \texttt{[Thought]} step. Examples are logical errors (\eg for an AND relation both statements need to be true, not just one), spatial reasoning errors (\eg confusing an overlap relation with left-of relation), and wanting to call the wrong tool (\eg use DetectObject instead of GoogleLens when the question requires recognizing a fine-grained entity).

{\emph{Tool call formulation}}. The orchestrator LLM makes a mistake when calling a tool in the \texttt{[Act]} step. Examples are syntax errors, wrong function arguments, or trying to use non-existing variables or non-existing tools.

{\emph{Tool failure}}. The tool fails to produce a correct answer given its input. For example, an object detector may fail to find objects of the input class, or the captioning tool may misinterpret the scene.

For our error analysis we inspect $10$ wrongly answered samples for each approach and for each question type (240 samples total).
We skip instances where the original question was ambiguous (\eg partial visibility makes exact counting hard).
If a \hammr agent was used as tool and returned the wrong result, we do not count it as a tool failure, but instead we dive into the agent itself to identify the root cause of the failure.
Examples can be found in Appendix~\ref{app:sec:additional_examples}.

\setlength{\tabcolsep}{6pt}
\begin{table}[t!]
    \caption{\textbf{Error analysis.} Percentage of mistakes per error type.}
    \label{tab:error_analysis}
    \centering
    \resizebox{\linewidth}{!}{
    \begin{tabular}{lcc|c}
         \toprule
         {\bf Error type} &  {\bf Naive Generic ReAct} & {\bf \hammr} & {\bf Specialist Agents} \\
         \midrule
         Reasoning              & 31.9\% & 20.9\% & 11.4\%\\
         Tool call formulation  & 20.6\% &  3.0\% &  5.0\%\\
         Tool failure           & 23.5\% & 35.8\% & 40.7\%\\
         \midrule                                  
         No error               & 24.0\% & 40.3\% & 42.9\%\\
         \bottomrule
    \end{tabular}
    }
    \vspace{-.3cm}
\end{table}

\para{Results.}
As shown in Tab.~\ref{tab:error_analysis} the naive generic agent makes many tool call formulation mistakes (20.9\%), whereas the specialist agents (5.0\%) and HAMMR (3.0\%) make rather few.
The majority of tool call formulation mistakes made by the generic agent are caused by calls to non-existing tools (\eg ColorOfObject, CountObjects). A few mistakes are caused by using non-existing variables, while syntax errors or using wrong function arguments of existing tools are rare.
Overall, this suggests that the large prompt of the naive generic ReAct agent confuses the LLM orchestrator about which tools it has at its disposal, and so it starts hallucinating them.

On the reasoning side, we find that the naive generic ReAct agent makes the most errors, followed by HAMMR and the specialists.
However, the type of reasoning errors made by different agents is very different.
For HAMMR, while many reasoning mistakes are made by individual specialist agents,
the additional reasoning errors are made by the Question Dispatcher agent.
The Question Dispatcher has an overall accuracy of 87\% and most of its errors are confusing two hop Encyclopedic VQA questions for single hop questions, and TextVQA questions for counting questions (if they are about a number) or for Encyclopedic VQA questions (if they are about detailed properties of an object).
In contrast, the naive generic agent often calls the wrong tool for the job, and makes assumptions without ever inspecting the image (\ie the first Thought \emph{before}
inspecting wthe image
is assuming that the object of interest is not present, or assuming that the plant/tree in the question refers to a cactus or sycamore).
Finally, we note that the generic ReAct agent has fewer tool failures than HAMMR and the specialist agents, which reflects that generic ReAct makes an error before it can call the correct tool.

\para{Can we improve the agents?}
The naive generic agent frequently hallucinates non-existing tools and makes many reasoning mistakes. Since it has a very long prompt where the interactions between all tools and in-context examples are unclear, it is difficult to improve this agent for the generic VQA setting.
In contrast, the additional errors made by \hammr over the upper-bound of the specialist agents are primarily caused by the Question Dispatcher agent.
Any improvements in the Question Dispatcher will immediately result in an overall improvement of the system (without improving the specialist agents).
Moreover, any improvement to any individual \hammr agents or their tools will improve the system overall.
Hence our compositional design brings multiple clear avenues for improving performance.
    
\vspace{-0.5em}
\subsection{Comparison to SOTA VLMs}
\label{sec:results_palix}

\setlength{\tabcolsep}{6pt}
\begin{table*}[t!]
\vspace{-.2cm}
    \caption{
    \textbf{Comparison to generic VLMs.}
    We compare HAMMR to several generic VLMs: BLIP-2~\cite{li23iclm_blip2}, Gemini Pro 1.0~\cite{gemini23arxiv}, and PaLI-X~\cite{chen23arxiv_palix}.
    HAMMR outperforms PaLI-X by $5.0\%$ on average, showing that it is a strong approach for generic VQA.
    }
    \label{tab:results_palix}
    \centering
    \resizebox{0.9\linewidth}{!}{
    \begin{tabular}{l|cc|ccc}
        \toprule
        Dataset        & BLIP-2~\cite{li23iclm_blip2} & HAMMR-BLIP-2 (ours) & Gemini Pro 1.0~\cite{gemini23arxiv} & PaLI-X \cite{chen23arxiv_palix} & HAMMR-PaLI-X (ours) \\
        \midrule
        PointQA local      & 25.3  & 47.8      & 16.2       & 48.2        & {\bf 69.1} \\
        PointQA look twice & 54.0  & 55.0      & 59.3       & 54.8        & {\bf 59.5} \\
        EncVQA single hop  & 15.3  & 45.0      & 22.1       & 18.5        & {\bf 47.8} \\
        EncVQA two hop     & 14.3  & \bf{22.8} & 16.6       & 10.2        & 22.0 \\
        NLVR2              & 52.2  & 55.4      & {\bf 70.5} & 64.8        & 63.3 \\
        GQA                & 52.9  & 50.5      & 61.3       & {\bf 76.4}  & 72.7 \\ 
        TallyQA            & 25.1  & 29.2      & 50.5       & {\bf 73.2}  & 72.0 \\
        TextVQA            & 26.3  & 16.3      & 72.0       & {\bf 70.1}  & 49.4 \\
        \midrule 
        Average performance & 33.2 & 40.3      & 46.1       & 52.0        & \bf{57.0} \\
        \bottomrule
    \end{tabular}
    }  %
\end{table*}

We compare \hammr against several modern generic Vision+Language Models~\cite{chen23arxiv_palix,gemini23arxiv,li23iclm_blip2} (VLMs).
Importantly, these models were trained on a large variety of tasks and report emerging capabilities on tasks not present in their training mix.
In particular, we compare to BLIP-2~\cite{li23iclm_blip2} (FlanT5-XXL) and Gemini Pro 1.0 Multimodal~\cite{gemini23arxiv}, both of which were trained on various tasks including VQA.
We also compare to PaLI-X~\cite{chen23arxiv_palix} (55B parameter version), which, after pre-training on a large variety of tasks, was fine-tuned on VQA tasks only (see Table 21 in~\cite{chen23arxiv_palix}).

In contrast to Sec.~\ref{sec:results_llm_tools}, where we wanted to highlight the ability of adapting existing models to new tasks, here we are primarily interested in achieving the best overall accuracy for generic VQA. We therefore instantiate a version of HAMMR that uses the more powerful PaLI-X~\cite{chen23arxiv_palix} VLM instead of BLIP-2 as a VQA tool.

The results in Tab.~\ref{tab:results_palix} reveal that \hammr outperforms PaLI-X by $5.0\%$ on average across all datasets, making it a strong approach for the generic VQA problem.
PaLI-X only outperforms \hammr on the datasets it was trained on, whereas \hammr is superior on the other datasets (on NLVR2 they perform comparably).
In line with previous work \cite{mensink23iccv_encvqa}, we observe especially low performance of the generic VLMs on encyclopedic questions (EncVQA rows).
These questions require accessing or memorizing detailed properties of fine-grained entities from a vast knowledge base. Such questions are difficult to answer with a classic VLM as it does not incorporate a retrieval mechanism.
In contrast, \hammr can naturally incorporate such capabilities by leveraging relevant tools (in our implementation: GoogleLens to recognize the entity and then Wikipedia to retrieve its page).

BLIP-2 and Gemini Pro Multimodal perform on average quite a bit lower than HAMMR and PaLI-X. 
This is likely because, in contrast to PALI-X, these models are not specialized for VQA by fine-tuning exclusively on it.
The more generalist PaLI-X model version also performs significantly worse on VQA than the specialized VQA model we use~\cite{chen23arxiv_palix}.
Finally, we note that BLIP-2 is a rather small model. 

Overall, this experiment demonstrates that \hammr can be easily adapted to leverage specialized VLMs, in addition to other tools, resulting in even higher performance (57.0 vs 40.3 in Tab.~\ref{tab:results_llm_tools}).
Hence, \hammr can leverage the best of both worlds: the wealth of implicit knowledge stored in a VLM, and the complementary knowledge that can be accessed by explicitly controlling tools.
    
\section{Conclusions}

We introduced \hammr: HierArchical MultiModal React, an evolution of the LLM+tools approach capable of tackling generic VQA.
We start from a multimodal ReAct-based system and make it hierarchical by enabling our \hammr agents to call upon other specialized agents. This enhances the compositionality of the LLM+tools approach which makes agents easier to develop and debug.
We demonstrate that the hierarchical agent setup is critical for obtaining high accuracy on generic VQA -- a BLIP-2 based version of \hammr outperforms a naive generic ReAct agent by $16.3\%$, while the improved HAMMR-PaLI-X version outperforms the strong generic PaLI-X VQA model by 5.0\%, achieving state-of-the-art results on our generic VQA suite.

{
    \small
    \bibliographystyle{IEEEtran}
    \bibliography{shortstrings,loco,loco_extra}
}

\clearpage
\setcounter{page}{1}

\appendix
\maketitle

\counterwithin{figure}{section}
\counterwithin{table}{section}
\setcounter{figure}{0}
\setcounter{table}{0}

\section{Additional examples}
\label{app:sec:additional_examples}

In this section we provide additional samples from the Naive generic ReAct agent and \hammr.
We include direct comparisons of the two systems on the same questions.
We additionally show reasoning errors from the QuestionDispatcherAgent, as well as one example of a tool call failure from the Naive generic ReAct agent as mentioned in the main paper.

\begin{figure*}[b!]
    \vspace{-.5cm}
    \centering
    \includegraphics[width=0.85\linewidth]{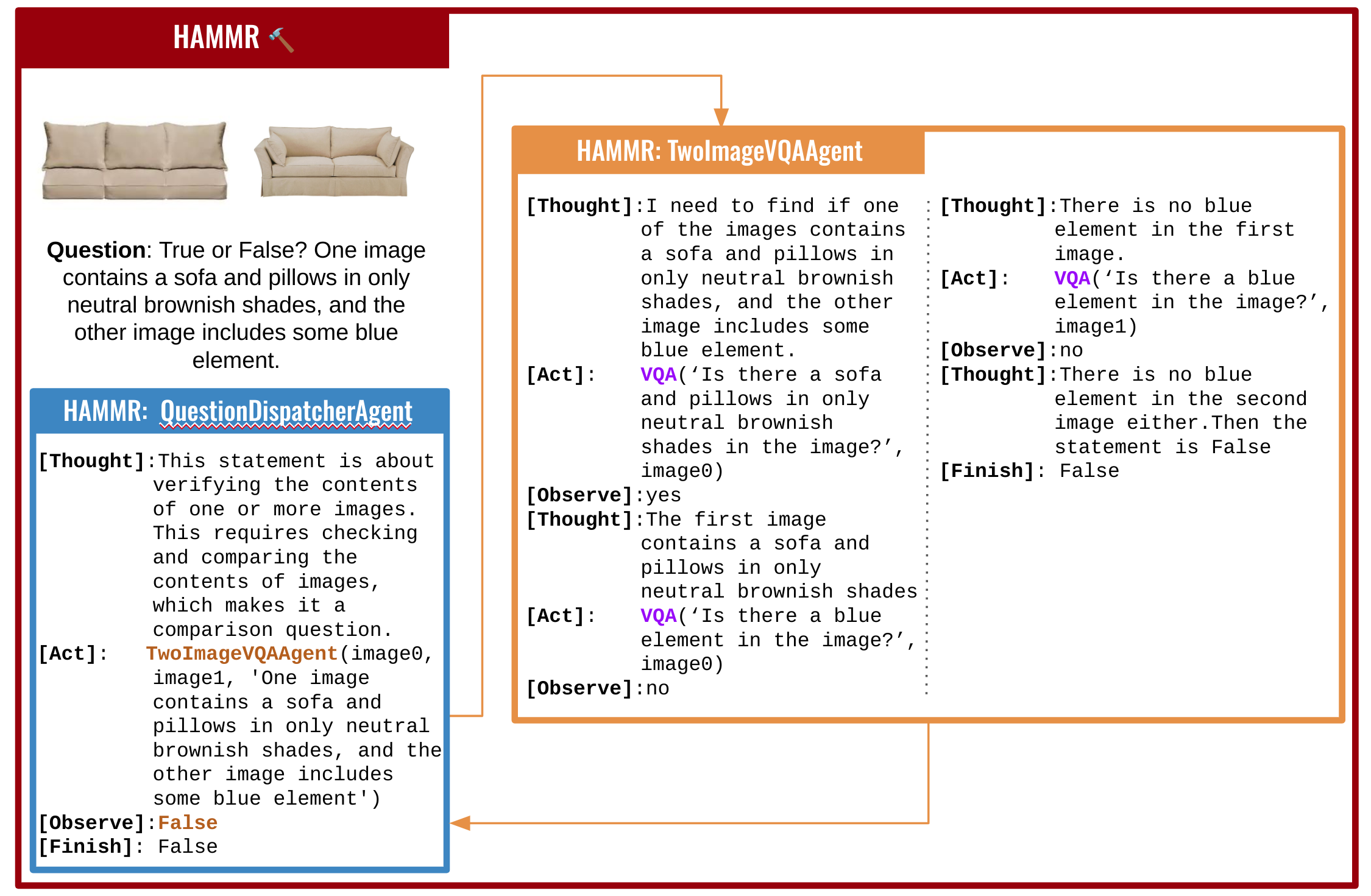}
    \caption{{\bf Example of \hammr answering a question that compares two images.}
    The \hammr QuestionDispatcherAgent first identifies that the question requires verifying and comparing properties of two images, and routes it to the appropriate agent. This agent starts by verifying the first statement - one image contains a sofa and pillows in only neutral brownish shades, which the first image fulfills. Then it checks the second property - the other image includes some blue element - on both images, which neither of them fulfills and therefore determines that the statement cannot be true. 
    }
\end{figure*}

\begin{figure*}[t!]
 \vspace{-.5cm}
    \centering
    \includegraphics[width=0.7\linewidth]{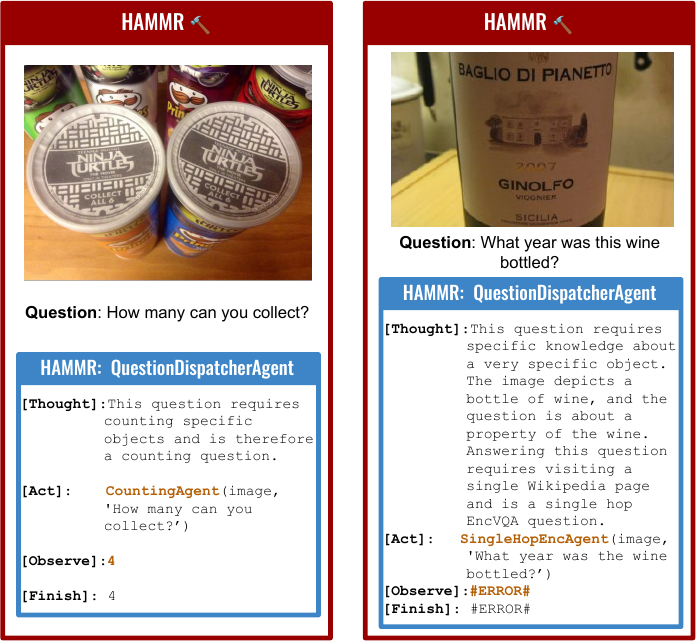}
    \caption{{\bf Examples of reasoning errors from the \hammr QuestionDispatcherAgent.}
    Both of these questions require the OCRReasoningAgent, but the HAMMR QuestionDispatcherAgent calls other agents to answer them. The agent interprets the first question as a counting question, although the information on how many cans can be collected is printed on them.
    For the second question, the answer can be found in the bottle label, but the agent identifies it as an encyclopedic question.
    }
\end{figure*}

\clearpage

\centering
\begin{tcolorbox}[title={\fontfamily{qhv} \center \textbf{Naive Generic ReAct}}, arc=0pt, left=0pt, right=0pt, bottom=0pt, top=0pt, boxrule=1.5pt, text width=0.95\linewidth, grow to left by=-3cm]

\begin{center}
    \includegraphics[width=.75\columnwidth]{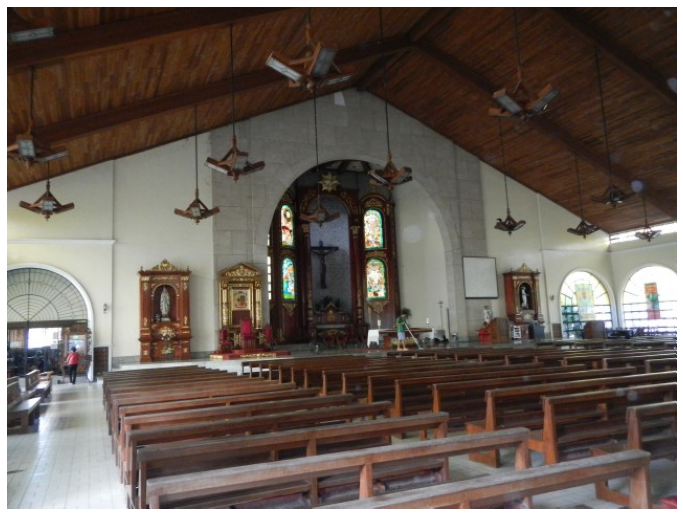}
    \rule{.9\columnwidth}{.5pt}
    \vspace{-5pt}
\end{center}

\scriptsize
\textbf{[Question]:} When was this cathedral dedicated to st. dominic de guzman?

\textbf{[Thought]:} I need to find the cathedral in the image.

\textbf{[Act]:} DetectObject(image, 'cathedral')

\textbf{[Observe]:} cathedral at left:250/top:161/width:120/height:161 (low confidence), cathedral at left:194/top:89/width:251/height:235 (low confidence), cathedral at left:0/top:8/width:632/height:463 (low confidence)

\textbf{[Thought]:} There are three potential cathedrals but they are all low confidence. Let's verify with vqa.

\textbf{[Act]:} ObjectInImage('cathedral', image)

\textbf{[Observe]:} yes

\textbf{[Thought]:} It was confirmed that there is a cathedral in the image. Now I need to find out when it was dedicated to st. dominic de guzman.

\textbf{[Act]:} VQA('When was this cathedral dedicated to st. dominic de guzman?', image)

\textbf{[Observe]:} 1990

\textbf{[Finish]:} 1990

\end{tcolorbox}

\noindent\begin{minipage}{\textwidth}
\vspace{-500pt}
\captionof{figure}{\textbf{Comparison between the Naive generic ReAct agent and \hammr.} The Naive generic ReAct agent uses PaLI to answer an encyclopedic question, which leads to a wrong answer. Instead, \hammr uses Lens to identify the cathedral in the image and finds the relevant information in the Wikipedia page of the building.}
\end{minipage}

\begin{tcolorbox}[title={\fontfamily{qhv} \center \textbf{HAMMR}}, arc=1pt, left=0pt, right=0pt, bottom=0pt, top=0pt, boxrule=1.5pt, colframe=purple, text width=0.95\linewidth, grow to right by=-3cm]

\begin{center}
    \includegraphics[width=.75\columnwidth]{figures/supp_images/generic_comparison_figure2.png}
    \rule{.9\columnwidth}{.5pt}
    \vspace{-5pt}
\end{center}

\begin{tcolorbox}[title={\fontfamily{qhv} \center \textbf{QuestionDispatcherAgent}},arc=0pt,left=0pt, right=0pt, bottom=0pt, top=0pt, boxrule=1.5pt, colframe=cyan]
\scriptsize
\textbf{[Question]:} When was this cathedral dedicated to st. dominic de guzman?

\textbf{[Thought]:} This question is about the dedication date of a cathedral. It requires external knowledge about the cathedral. It is an EncyclopedicSingleHop question.

\textbf{[Act]:} {\color{magenta} \textbf{EncyclopedicSingleHopAgent}}( 'When was this cathedral dedicated to st. dominic de guzman?', image)

\textbf{[Observe]:} September 8, 1739

\textbf{[Finish]:} September 8, 1739

\end{tcolorbox}

\begin{tcolorbox}[title={\fontfamily{qhv} \center \textbf{SingleHopEncyclopedicAgent}},arc=0pt,left=0pt, right=0pt, bottom=0pt, top=0pt, boxrule=1.5pt, colframe=magenta]
\scriptsize
\textbf{[Question]:} When was this cathedral dedicated to st. dominic de guzman?

\textbf{[Thought]:} First I need to identify the cathedral in the image.

\textbf{[Act]:} GoogleLens(image)

\textbf{[Observe]:} Bayombong Cathedral

\textbf{[Thought]:} Now I need to extract the article from the most confident entity.

\textbf{[Act]:} article = WikipediaArticle("Bayombong Cathedral")

\textbf{[Observe]:} I have stored the output of WikipediaArticle in the following variables: article.

\textbf{[Thought]:} Now I need to answer the question according to the article.

\textbf{[Act]:} AnswerWithContext('''When was this cathedral dedicated to st. dominic de guzman?''', article)

\textbf{[Observe]:} September 8, 1739

\textbf{[Finish]:} September 8, 1739

\end{tcolorbox}
\end{tcolorbox}

\clearpage
\begin{tcolorbox}[title={\fontfamily{qhv} \center \textbf{Naive Generic ReAct}},arc=0pt,left=0pt,
right=0pt, bottom=0pt, top=0pt, boxrule=1.5pt, text width=0.95\columnwidth, grow to left by=-3cm]

\begin{center}
    \includegraphics[width=.75\columnwidth]{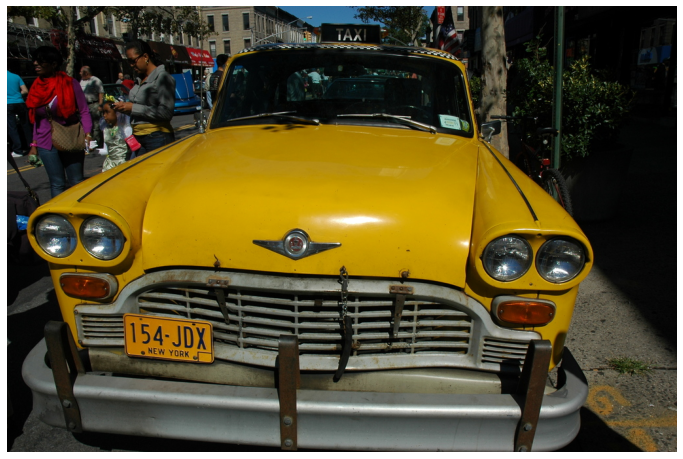}
    \rule{.9\columnwidth}{.5pt}
    \vspace{-5pt}
\end{center}

\scriptsize

\textbf{[Question]:} what is the license plate?

\textbf{[Thought]:} The license plate is not in the image.

\textbf{[Finish]:} no

\end{tcolorbox}

\vspace{350pt}

\noindent\begin{minipage}{\textwidth}
\centering
\vspace{-15.5cm}
\captionof{figure}{\textbf{Comparison between the Naive generic ReAct agent and \hammr.} The Naive generic ReAct assumes that the license plate is not in the image and provides a wrong answer. \hammr correctly identifies the OCR-based reasoning question, then it detects the license plate, and finally it performs OCR on the license plate crop.}
\end{minipage}

\begin{tcolorbox}[title={\fontfamily{qhv} \center \textbf{HAMMR}}, arc=0pt,left=0pt, right=0pt, bottom=0pt, top=0pt, boxrule=1.5pt, colframe=purple, text width=0.95\columnwidth, grow to right by=-3cm]

\begin{center}
    \includegraphics[width=.75\columnwidth]{figures/supp_images/generic_comparison_figure3.png}
    \rule{.9\columnwidth}{.5pt}
    \vspace{-5pt}
\end{center}

\begin{tcolorbox}[title={\fontfamily{qhv} \center \textbf{QuestionDispatcherAgent}},arc=0pt,left=1pt, right=1.5pt, bottom=1pt, top=1pt, boxrule=2pt, colframe=cyan]

\scriptsize
\textbf{[Question]:} what is the license plate?

\textbf{[Thought]:} Let's caption the image.

\textbf{[Act]:} Caption(image)

\textbf{[Observe]:} a yellow taxi cab parked on the side of the road .

\textbf{[Thought]:} To answer the question I need to read the license plate. This does not require external or encyclopedic knowledge, it is a non-encyclopedic question requiring OCR-based reasoning.

\textbf{[Act]:} {\color{violet} \textbf{OCRReasoningAgent}}('what is the license plate?', image)

\textbf{[Observe]:} 154-jdx new york

\textbf{[Finish]:} 154-jdx new york

\end{tcolorbox}

\begin{tcolorbox}[title={\fontfamily{qhv} \center \textbf{OCRReasoningAgent}},arc=0pt,left=0pt, right=0pt, bottom=0pt, top=0pt, boxrule=1.5pt, colframe=violet]
\scriptsize
\textbf{[Question]:} what is the license plate?

\textbf{[Thought]:} I need to find a license plate in the image.

\textbf{[Act]:} DetectObject(image, 'license plate')

\textbf{[Observe]:} license plate at left:174/top:466/width:141/height:80, license plate at left:12/top:317/width:21/height:26, license plate at left:659/top:164/width:47/height:28

\textbf{[Thought]:} I found three potential license plates. I need to crop them.

\textbf{[Act]:} crops = CropImage(image, [[174, 466, 141, 80], [12, 317, 21, 26], [659, 164, 47, 28]])

\textbf{[Thought]:} I need to OCR the content of the license plates.

\textbf{[Act]:} OCR(crops)

\textbf{[Observe]:} [154-jdx new york, cannot answer - the image quality is too poor, cannot answer - the image quality is too poor].

\textbf{[Finish]:} 154-jdx new york

\end{tcolorbox}
\end{tcolorbox}

\clearpage
\begin{tcolorbox}[title={\fontfamily{qhv} \center \textbf{Naive Generic ReAct}},
arc=0pt,left=0pt, right=0pt, bottom=0pt, top=0pt, boxrule=1.5pt, text width=0.65\columnwidth]

\begin{center}
    \includegraphics[width=.85\columnwidth]{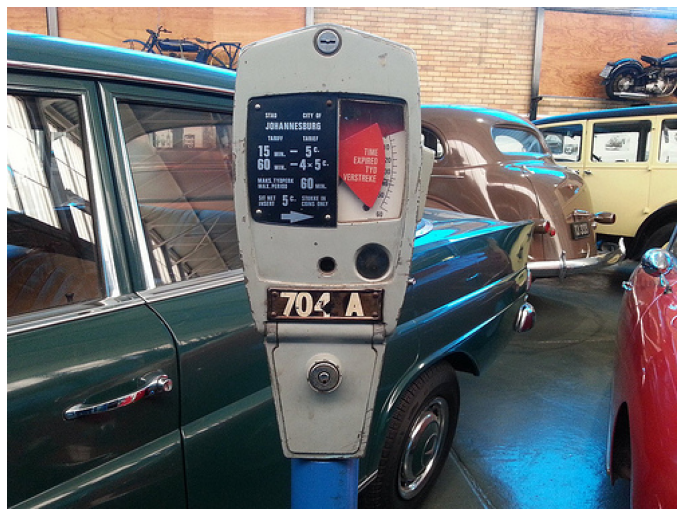}
    \rule{.9\columnwidth}{.5pt}
    \vspace{-5pt}
\end{center}

\scriptsize

\textbf{[Question]:} How many yellow cars are in the picture?

\textbf{[Thought]:} I need to find the cars in the image.

\textbf{[Act]:} DetectObject(image, 'car')

\textbf{[Observe]:} car at left:447/top:153/width:52/height:218, car at left:304/top:69/width:133/height:139, car at left:0/top:14/width:424/height:360, car at left:388/top:69/width:111/height:122

\textbf{[Thought]:} I need to find the yellow cars.

\textbf{[Act]:} FilterObjects(image, 'car', 'color', 'yellow')

\textbf{[Observe]:} \#ERROR\#: Tool FilterObjects is not registered.

\textbf{[Finish]:} \#ERROR\#: Tool FilterObjects is not registered.

\end{tcolorbox}

\noindent\begin{minipage}{\columnwidth}
\captionof{figure}{\textbf{Tool call failure in the Naive generic ReAct agent.} The Naive generic ReAct hallucinates a \texttt{FilterObjects} tool, and it is incapable of recovering from this error.}
\end{minipage}

\end{document}